\documentclass[a4paper]{article}

\usepackage{INTERSPEECH2022}
\usepackage{multirow}
\usepackage{comment}
 
\usepackage{color, colortbl, cite}
\usepackage{algorithm}
\usepackage{algorithmic}
\usepackage{arydshln}

\title{InterAug: Augmenting Noisy Intermediate Predictions for CTC-based ASR}

\name{Yu Nakagome, Tatsuya Komatsu, Yusuke Fujita, Shuta Ichimura, Yusuke Kida}
\address{LINE Corporation}
\email{nakagome.yu@linecorp.com}

\begin{document}

\maketitle
\begin{abstract}
This paper proposes InterAug: a novel training method for CTC-based ASR using augmented intermediate representations for conditioning.
The proposed method exploits the conditioning framework of self-conditioned CTC to train robust models by conditioning with ``noisy'' intermediate predictions.
During the training, intermediate predictions are changed to incorrect intermediate predictions, 
and fed into the next layer for conditioning.
The subsequent layers are trained to correct the incorrect intermediate predictions with the intermediate losses.
By repeating the augmentation and the correction, iterative refinements, which generally require a special decoder, can be realized only with the audio encoder.
To produce noisy intermediate predictions, 
we also introduce new augmentation: intermediate feature space augmentation and intermediate token space augmentation that are designed to simulate typical errors.
The combination of the proposed InterAug framework with new augmentation allows explicit training of the robust audio encoders.
In experiments using augmentations simulating deletion, insertion, and substitution error, 
we confirmed that the trained model acquires robustness to each error, 
boosting the speech recognition performance of the strong self-conditioned CTC baseline.

\end{abstract}


\section{Introduction}
With the significant development of deep neural networks, end-to-end (E2E) automatic speech recognition (ASR) systems~\cite{hinton2012deep, graves2013speech} achieve tremendous performance and become the mainstream of ASR research.
E2E models can be generally divided into an autoregressive (AR) ASR model~\cite{graves2012sequence, bahdanau2014neural} 
and non-autoregressive (NAR) ASR~\cite{Graves06_icml, gu2017non} models.
On the one hand, AR models consist of an encoder that encodes acoustic features and a decoder that predict sentences.
AR models have shown state-of-the-art recognition performance, they output tokens one by one, so that each token depends on all previously generated tokens, increasing latency.
NAR models, on the other hand, have attracted interest because of their ability to perform efficient inference, predicting all tokens simultaneously.
They are faster for decoding, and lighter in size than AR models.
In addition, although it has the restriction of assuming conditional independence among output tokens, recent improved methods~\cite{Chen21_SPL, Higuchi20b_interspeech, Chi21_NAACL, lee21_icassp, nozaki21_interspeech, chan2020imputer, bai2020listen, fujita2020insertion, tian2020spike}, describe below,
perform close to the recognition accuracy of the AR methods. 
In recent years, there has been a growing demand for technologies that allow ASR systems to operate on devices such as smartphones and tablets.
Since these devices generally have limited computational resources, fast and lightweight inference processing is a key element of on-device ASR. Therefore, improving the accuracy of NAR models is an important field of research.

NAR ASR approaches can be generally divided into iterative refinement decoding \cite{Chen21_SPL, Higuchi20b_interspeech, Chi21_NAACL, chen2021align} and intermediate prediction objective \cite{lee21_icassp, nozaki21_interspeech}.
Iterative refinement decoding consists of an encoder and a NAR decoder, where the encoder is trained with the CTC objective, and the decoder is trained with some refinement model objectives.
In Mask-CTC~\cite{Higuchi20b_interspeech} and Improved Mask-CTC~\cite{higuchi2021improved}, low-confidence tokens in the CTC output are masked before the decoding stage.
These tokens are iteratively refined by using the masked language model conditioned on the other unmasked tokens.
Align-Refine~\cite{Chi21_NAACL} and Align-Denoise~\cite{chen2021align} input a latent alignment of CTC to the decoder and refine them over the alignment space.
In these methods, an initial prediction from the encoder can be iteratively refined by editing low-confidence tokens with the trained refinement model.
These iterative refinement decoding approaches make training of NAR models more tractable by defining objective for errors in the output token space, showing promising improvements.

The models with intermediate prediction objectives consist of an audio encoder with a transformer/conformer~\cite{vaswani2017attention, gulati20_interspeech} and a CTC decoder.
Intermediate CTC~\cite{lee21_icassp} and Self-conditioned CTC~\cite{nozaki21_interspeech} outperform many NAR methods even though they do not use a special decoder.
Intermediate CTC makes a prediction of the output sequence at the intermediate layer and takes the CTC loss as well as the final layer.
Self-conditioned CTC extends Intermediate CTC by adding intermediate predictions to the input of the subsequent layers.
While Intermediate CTC and Self-conditioned CTC have excellent recognition performance by boosting the capability of their audio encoders, 
the CTC loss, which is defined over sequences, is not easy to handle token-level objective like iterative refinement approaches.

%

In this study, we attempt to incorporate the essence of the iterative refinement model into training CTC-based models using self-conditioning framework and `noisy' data conditioning.
We proposes to intentionally corrupt the token predictions at an intermediate layer by deletions, insertions, and substitutions.
Then, these augmented predictions are used as conditioning features fed into the next encoder layer.
By correcting these errors through the multiple intermediate augmentations and subsequent intermediate losses,
a similar operation with iterative refinement can be realized within the single audio encoder.
This allows the proposed method to incorporate the advantage of iterative refinement decoding and intermediate prediction objective without requiring a special decoder and keeping the model size lightweight.
In addition to conditioning the corrupted intermediate tokens, augmentations such as time-masking and feature-masking in the feature space are also proposed.
The augmented features are similarly used as a condition to the input of the later encoder layer, leading to improved robustness of the model.



\begin{figure*}[!ht]
 \begin{center}
 \includegraphics[width=2.0\columnwidth]{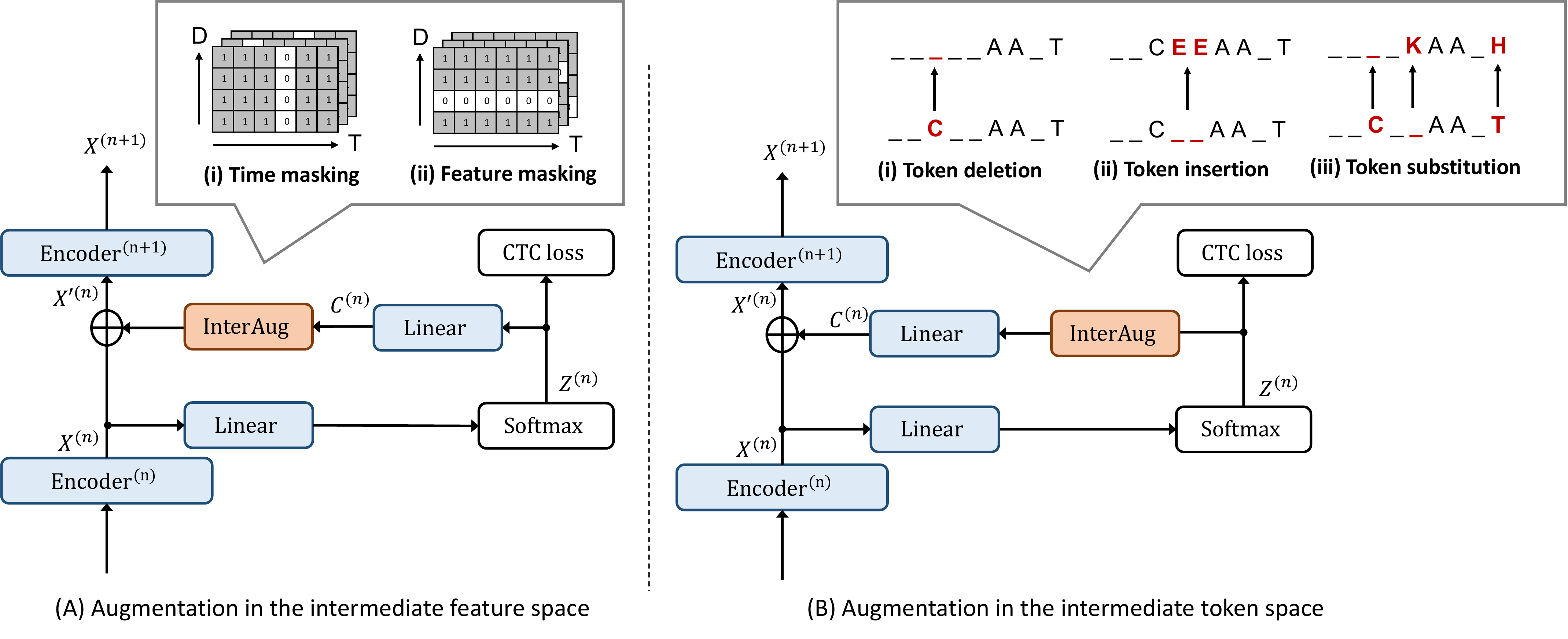}
 \end{center}
 \caption{Overview of the proposed augmentation methods. The proposed method has two types: (A) Augmentation such as time masking and feature masking is applied to intermediate feature. (B) Deletion, Insertion, Substitution are applied to intermediate tokens. These augmented errors are corrected by subsequent encoders, which contributes to enhance the robustness of the model.}
 \label{fig:prop-diagram}
\end{figure*}

\section{Self-conditioned CTC}
\label{sec:sc-ctc}
In this section, we give an overview of CTC-based ASR and Self-conditioned CTC \cite{nozaki21_interspeech} which is the backbone of the proposed InterAug framework.
\subsection{Connectionist Temporal Classification}
\label{sec:ctc}
End-to-end ASR aims to model the probability distribution of a token sequence $Y=(y_l \in \mathcal{V} \mid l=1,\dots,L)$ given a sequence of $D$-dimensional audio features $X=(\mathbf{x}_t \in \mathbb{R}^D \mid t=1,\dots,T)$, where $\mathcal{V}$ is a token vocabulary.
In the CTC framework \cite{Graves06_icml}, frame-level alignment paths between $X$ and $Y$ is introduced with a special blank token $\epsilon$.
An alignment path is denoted by $\pi=\left(\pi_t \in \mathcal{V}' \mid t=1,\dots,T\right)$, where $\mathcal{V}'= \mathcal{V} \cup \{ \epsilon \}$.
The alignment path can be transformed into the corresponding token sequence by using the collapsing function $\mathcal{B}$ that removes all repeated tokens and blank tokens.
A neural network is trained to estimate the probability distribution of $\pi_t$.
An output sequence of the neural network is denoted by $Z = (\mathbf{z}_t \in (0,1)^{|\mathcal{V}'|} \mid t=1,\dots,T)$, where the element $z_{t,k}$ is interpreted as $p(\pi_t = k |X)$.
The training objective of CTC is the negative log-likelihood over all possible alignment paths with the conditional independence assumption per frame, as follows:
\begin{align}\label{eq:lossctc}
    \mathcal{L}_\mathsf{ctc}(Z, Y) = - \log \sum_{\pi\in\mathcal{B}^{-1}(Y)}\prod_{t} z_{t,\pi_t}.
\end{align}

\subsection{Conformer Encoder}
\label{sec:conformer}
The neural network used in this paper has $N$-stacked Conformer encoders \cite{gulati20_interspeech}.
The $n$-th encoder accepts an input sequence $X^{(n-1)}$ and produces an encoded sequence of the same shape:
\begin{align}
    X^{(n)} = \mathsf{Encoder}^{(n)}(X^{(n-1)}) \qquad (1 \le n \le N),
\end{align}
where $X^{(0)} = X$ is a subsampled sequence of input audio features.
The output sequence $Z$ is obtained by applying a linear transformation and the softmax function:
\begin{align}
\label{eq:out}
    Z = \mathsf{Softmax}(\mathsf{Linear}_{D\rightarrow |\mathcal{V}'|}(X^{(N)})),
\end{align}
where $\mathsf{Linear}_{D\rightarrow |\mathcal{V}'|}(\cdot)$ maps a $D$-dimensional vector into  a $|\mathcal{V}'|$-dimensional vector for each element of $X^{(N)}$.

\subsection{Intermediate CTC and Self-conditioned CTC}
For regularizing the CTC model training, Intermediate CTC \cite{lee21_icassp} introduces an additional loss for output sequences of intermediate encoders.
An intermediate output sequence for the $n$-th encoder $Z^{(n)} = (\mathbf{z}^{(n)}_t \in (0,1)^{|\mathcal{V}'|}| t=1,\dots,T)$ is computed using the same linear transformation as Eq.~\ref{eq:out}:
\begin{align}
\label{eq:softmax}
    Z^{(n)} = \mathsf{Softmax}(\mathsf{Linear}_{D\rightarrow |\mathcal{V}'|}(X^{(n)})).
\end{align}
The loss for the intermediate output sequence is the same as Eq.~\ref{eq:lossctc}, and is added to the final training objective as follows:
\begin{equation}
    \mathcal{L}_\mathsf{ic} = (1-\lambda)\mathcal{L}_\mathsf{ctc}(Z,Y) + \frac{\lambda}{|\mathcal{N}|}\sum_{n \in \mathcal{N}} \mathcal{L}_\mathsf{ctc}(Z^{(n)},Y),
\end{equation}
where $\lambda \in (0,1)$ is a mixing weight and $\mathcal{N}$ is a set of encoder indices for intermediate losses.

Self-conditioned CTC \cite{nozaki21_interspeech} utilizes the intermediate output sequence for conditioning the subsequent encoders:
\begin{align}
    X^{(n)} &= \mathsf{Encoder}^{(n)}(X'^{(n-1)}) \qquad (1 \le n \le N), \label{eq:encoder-output}\\
    C^{(n)} &= \mathsf{Linear}_{|\mathcal{V}'|\rightarrow D}(Z^{(n)}), \label{eq:linear-prime}\\
    X'^{(n)} &=
        \begin{cases}
            X^{(n)} + C^{(n)} & (n \in \mathcal{N}), \\
            X^{(n)} & (n \notin \mathcal{N}),
        \end{cases}
\label{eq:selfcond}
\end{align}
 where $C^{(n)} = (\mathbf{c}^{(n)}_t \mid t=1,\dots,T)$ and $\mathsf{Linear}_{|\mathcal{V}'|\rightarrow D}(\cdot)$ maps a $|\mathcal{V}'|$-dimensional vector into a $D$-dimensional vector for each element in the input sequence.
This linear layer is shared among the intermediate layers.

\section{InterAug: Augmenting Noisy Intermediate Predictions}
\label{sec:interaug}
Figure~\ref{fig:prop-diagram} illustrates the proposed InterAug framework.
The proposed method changes the intermediate predictions into augmented noisy predictions~$C_\mathsf{Aug}$ and conditions them on the encoder output~$X^{(n)}$ in Eq.~\ref{eq:encoder-output} as follows:
\begin{align}
    X'^{(n)} = X^{(n)} + C_\mathsf{Aug}.
\end{align}
%
The intermediate losses at the subsequent layers correct the errors contained in the noisy predictions.
The repetition of augmentation and correction helps models to be robust to noise.
%
To simulate noisy predictions, 
we apply two types of augmentation for intermediate predictions as Fig~\ref{fig:prop-diagram}:
(A) augmentation in the intermediate feature space and 
(B) augmentation in the intermediate token space.
\subsection{Augmentation in the intermediate feature space}\label{aug_feat}
This augmentation is applied to the sequence in the feature space~$C^{(n)}$ in Eq.~\ref{eq:linear-prime} which is obtained by projected intermediate predictions~$Z^{(n)}$ in Eq.~\ref{eq:softmax} back to the space of the encoder feature as follows:
\begin{align}
    C_{\mathsf{Aug}} = \mathsf{FeatAug}(C^{(n)}),
\end{align}
where the function $\mathsf{FeatAug(\cdot)}$ represents an augmentation function in the feature space.
Inspired by SpecAugment~\cite{Park19_interspeech}, we apply masking blocks of time steps and feature channels for the augmentation with probability $p_{time}$ and $p_{feat}$, respectively.

\noindent\textbf{(i)~Time masking:}
Time masking is applied so that $\tau$ consecutive time frames $[ t_{0}, t_{0}+\tau ]$ are masked by replacing the elements with zeros.
$\tau$ is chosen from the uniform distribution $\mathcal{U}(0, W_\tau)$, and $t_{0}$ is chosen from $\mathcal{U}(0, T - \tau)$, where $W_\tau$ is a hyperparameter to control the masking duration.

\noindent\textbf{(ii)~Feature masking:}
Similarly, feature masking is applied so that $d$ consecutive feature channels $[d_{0}, d_{0}+d]$ are masked, where $d$ is chosen from $\mathcal{U}(0, W_d)$ and $d_{0}$ is chosen from $\mathcal{U}(0, D - d)$. $W_d$ is a hyperparameter to control the masking span for the feature channel dimension.

\subsection{Augmentation in the intermediate token space}
This approach performs augmentation on intermediate predictions $Z^{(n)}$ in Eq.~\ref{eq:softmax} as follows:
\begin{align}
    C_{\mathsf{Aug}} = \mathsf{Linear}_{|\mathcal{V}'|\rightarrow D}(\mathsf{TokenAug}(Z^{(n)})),
\end{align}
where the function $\mathsf{TokenAug(\cdot)}$ represents an augmentation function in the token space.
Since the intermediate predictions are in the same domain as the output,
it is possible to design augmentations simulating errors that occur upon prediction,
such as insertions, deletions, and substitution errors.
By simulating and correcting these errors through the multiple intermediate augmentations and subsequent intermediate losses,
the iterative refinement~\cite{Chi21_NAACL} like operation can be realized within the single audio encoder. 

This paper designs three types of augmentation simulating the actual errors: 
(1) Token deletion for deletion errors, (2) Token insertion for insertion errors, and (3) Token substitution for substitution errors.

\noindent\textbf{(i)~Token Deletion} simulates the deletion errors, which drop tokens to be predicted.
The deletion error can be simulated by randomly replacing estimated token label with the blank label (at index 0) as follows:

\begin{align}
    r^\mathsf{(del)}_t &\sim \mathsf{Bernoulli}(p_\mathsf{del}), \\
    \tilde{\pi}^{(n)}_t &= r^\mathsf{(del)}_t \arg\max_k z^{(n)}_{t,k}, \\
    \mathbf{c}^{(n)}_{t} &= \mathsf{Onehot}(\tilde{\pi}^{(n)}_t),
\end{align}
where $p_\mathsf{del}$ is the deletion probability.

\noindent\textbf{(ii)~Token Insertion} simulates the insertion errors, which predict unnecessary tokens where it should be blank tokens.
The token estimated as blank is replaced with the token that has the second highest posterior by randomly masking the posterior of the blank label:
\begin{align}
    r^\mathsf{(ins)}_t &\sim \mathsf{Bernoulli}(p_\mathsf{ins}), \\
    z'^{(n)}_{t,k} &= \begin{cases}
            -\infty & r^\mathsf{(ins)}_t = 1 \land k = 0, \\
            z^{(n)}_{t,k} & \mathsf{otherwise},
            \end{cases} \\
    \tilde{\pi}^{(n)}_t &= \arg\max_k z'^{(n)}_{k,t}, \\
    \mathbf{c}^{(n)}_{t} &= \mathsf{Onehot}(\tilde{\pi}^{(n)}_t),
\end{align}
where $p_\mathsf{ins}$ is the insertion probability.

\noindent\textbf{(iii)~Token Substitution} intentionally adds token substitutions to the indeterminate prediction.
The token is directly sampled from the posterior distribution $Z^{(n)}$ as a replacement:
\begin{align}
    \tilde{\pi}^{(n)}_t &\sim P(\pi_t) : P(\pi_t = k) = z^{(n)}_{t,k}, \\
    \mathbf{c}^{(n)}_{t} &= \mathsf{Onehot}(\tilde{\pi}^{(n)}_t).
\end{align}
It should be noted here that the token substitution intrinsically contains both deletion and insertion errors as well as substitution errors, since sampling may convert a token from blank to non-blank, or from non-blank to blank.


\section{Experiments}
\label{sec:experiment}
To evaluate the effectiveness of InterAug, we conducted speech recognition experiments using ESPnet~\cite{watanabe18_interspeech, Pengcheng21_icassp}.
The performance of the models was evaluated based on word error rates (WERs), word substitution error rates (sub), word deletion error rates (del) and word insertion error rates (ins).
All methods were evaluated by greedy decoding without relying on external language models.

\subsection{Data}
The experiments were conducted on 100 hours subset of LibriSpeech~\cite{Librispeech}.
LibriSpeech consists of utterances from read English audiobooks. 
For the network input samples, 80-dimensional Mel-scale filterbank coefficients with three-dimensional pitch features were extracted using Kaldi toolkit~\cite{Povey11_ASRU}.
Speed perturbation~\cite{Ko15_interspeech} and SpecAugment~\cite{Park19_interspeech} were also applied to the training data.
For the tokenization of texts, we created 300 subwords for LibriSpeech with SentencePiece~\cite{kudo-richardson-2018-sentencepiece}.


\subsection{Model configurations}

\textbf{CTC:}
We used the Conformer-CTC model as described in Section~\ref{sec:conformer}.
The number of layers $N$ was 18, and the encoder dimension $D$ was 256.
The convolution kernel size and the number of attention heads were 15 and 4, respectively.
The feed-forward layer dimension in the Conformer blocks was set to 1024.
The model was trained for 50 epochs, and the final model was obtained by averaging model parameters over 10-best checkpoints in terms of validation loss values.
The batch-size is set to 128 for LibriSpeech. 
The Adam optimizer~\cite{Kingma14_iclr} with $\beta_1 = 0.9$, $\beta_2=0.98$,
the Noam learning rate scheduling~\cite{Vaswani17_NIPS} with 25k warmup steps, and a learning rate factor of 5.0 were used for training.

\noindent\textbf{SelfCond:}
Five intermediate CTC predictions are applied at the set of layer indices $\mathcal{N} = \{3,6,9,12,15\}$ with $\lambda = 0.5$ and condition to the next layers according to Eq.~\ref{eq:selfcond}.

\subsection{InterAug configurations}
We applied InterAug based on the Self-conditioned CTC architecture with five augmentation approaches as described in Section~\ref{sec:interaug}:
1) Time masking and 2) Feature masking for the feature space augmentation, 
3) Token deletion, 4) Token insertion and 5) Token substitution for the token space augmentation.

Probability to apply each augmentation is  empirically decided based on preliminary experiments.
For feature space augmentation, $p_{time}$ and $p_{feat}$ in Section~\ref{aug_feat} was set to 1.0.
For token space augmentation, both $p_{del}$ and $p_{ins}$ for token deletion and insertion were set to 0.1.
Token substitution has no probability parameter, since it performs augmentation based on the obtained intermediate probability.
Note that during the inference phase, any augmentation is not applied.





\begin{table}[t]
\centering
\caption{
Word error rates (WERs) on LibriSpeech (100h). The results were obtained without language models.
}
\label{tab:main}
\begin{tabular}{lcccc}\hline
\multirow{2}{*}{\textbf{Model}} &
\multicolumn{2}{c}{\textbf{dev}} & \multicolumn{2}{c}{\textbf{test}} \\
 & \textbf{clean} & \textbf{other} & \textbf{clean} & \textbf{other} \\\hline
\multicolumn{4}{l}{{\bf Conventional models}}\\
CTC       & 8.69 & 23.00 & 9.08 & 23.68\\
SelfCond  & 7.11 & 20.82 & 7.48 & 21.31\\ \hdashline
\multicolumn{4}{l}{{\bf InterAug~(Feature space)}} \\
Time masking   & 6.97 & 20.47 & 7.44 & 20.87\\
Feature masking  & 8.42 & 23.46 & 8.79 & 23.79\\ \hdashline
\multicolumn{4}{l}{{\bf InterAug~(Token space)}} \\
Token deletion  & 6.97 & 20.30 & 7.38 & 20.72\\
Token insertion  & 6.95 & 20.32 & 7.44 & 20.71\\
Token substitution & \bf{6.90} & \bf{19.99} & \bf{7.23} & \bf{20.34}\\
\hline
\end{tabular}%
\vspace{5mm}
\end{table}

\begin{table}[t]
\centering
\caption{Error analysis on test-clean and test-other of LibriSpeech (100h).}
\label{tab:sub-del-ins}
\begin{tabular}{lcccccc}
\hline
\multirow{2}{*}{\textbf{Method}} & \multicolumn{3}{c}{\textbf{test-clean}} & \multicolumn{3}{c}{\textbf{test-other}} \\
 & \multicolumn{1}{c}{\textbf{sub}} & \multicolumn{1}{c}{\textbf{del}} & \multicolumn{1}{c}{\textbf{ins}} & \multicolumn{1}{c}{\textbf{sub}} & \multicolumn{1}{c}{\textbf{del}} & \multicolumn{1}{c}{\textbf{ins}} \\\hline
\multicolumn{1}{l}{{\bf Baseline}} \\
SelfCond & 6.07 & 0.65 & 0.77 & 17.09 & 2.27 & \bf{1.95}\\\hdashline
\multicolumn{4}{l}{{\bf InterAug~(Feature space)}} \\
Time mask & 6.06 & \bf{0.60} & 0.78 & 16.76 & 2.05 & 2.07	\\
Feat. mask & 7.04 & 0.63 & 1.12 & 18.95 & 2.08 & 2.76 \\ \hdashline
\multicolumn{4}{l}{{\bf InterAug~(Token space)}} \\
Token del. & 5.98 & \bf{0.60} & 0.80 & 16.70 & \bf{1.95} & 2.07 \\
Token ins. & 5.98 & 0.67 & \bf{0.74} & 16.46 & 2.08 & 1.97 \\
Token sub. & \bf{5.84} &	0.61 & 0.78 & \bf{16.34} & 1.99 & 2.00\\
\hline
 
\end{tabular}
\end{table}

\begin{table}[t]
\centering
\caption{Analysis for augmentation position on LibriSpeech (100h).
There are two candidates to apply the augmentation function $\mathsf{SA}(\cdot)$:
(1) Encoder feat.: $X'{(n)} = \mathsf{SA}(X^{(n)}) + C^{(n)} $, 
(2) Cond. feat.: $X'{(n)} = X^{(n)} + \mathsf{SA}(C^{(n)}) $.
Time-masking ($W_{\tau}=0.1T$, $p_{time}$=0.5) was applied to each feature.
}
\label{tab:pos-aug2}
\begin{tabular}{lcccc}
\hline
\multirow{2}{*}{\textbf{Augmentation position}} & \multicolumn{2}{c}{\textbf{dev}} & \multicolumn{2}{c}{\textbf{test}} \\
 & \multicolumn{1}{c}{\textbf{clean}} & \multicolumn{1}{c}{\textbf{other}} & \multicolumn{1}{c}{\textbf{clean}} & \multicolumn{1}{c}{\textbf{other}}\\\hline
\multicolumn{2}{l}{{\bf w/o. Augmentation}} \\
SelfCond  & 7.11 & 20.82 & 7.48 & 21.31\\ \hdashline
\multicolumn{1}{l}{{\bf w. Augmentation}} \\
(1)~Encoder feat. & 9.51 & 25.71 & 9.74 & 26.16\\ 
(2)~Cond. feat. (Prop.) & \textbf{6.97} & \textbf{20.47} & \textbf{7.44} & \textbf{20.87} \\
\hline
\end{tabular}
\end{table}

\subsection{Results}
\label{sec:results}
Table~\ref{tab:main} summarizes the experimental results.
First, it can be seen that the proposed time masking, token deletion, token insertion, and token substitution boost the Self-conditioned CTC on all evaluation sets.
In particular, it shows that token substitution perform the best recognition result among the methods compared.

A more detailed error analysis on test subsets of LibriSpeech (100h) is presented in Table~\ref{tab:sub-del-ins}.
Compared to Self-conditioned CTC, time masking and token deletion were confirmed to improve the deletion error rate.
In addition, token insertion slightly reduced insertion errors in the test-clean subset.
Moreover, token substitution significantly reduced substitution errors and deletion errors.
The reason that token substitution performed best was that the CTC prediction series usually contained many blank tokens, so substitution from blanks to non-blank tokens was often performed, which had the combined effect of correcting not only substitution errors but also deletion errors.
Feature masking was found to be ineffective since the recognition accuracy degraded in Table 1.
Based on these results, the proposed method can train models with desired robustness to deletion, substitution, and insertion errors using corresponding intermediate augmentation.
Using the proposed InterAug framework, we can realize more tractable training of the audio encoders.


\subsection{Analysis for Augmentation Position}
\label{sec:result-aug-pos}
In this section, we report an additional study on the effectiveness of augmenting the the conditioning features $C^{(n)}$ rather than the encoder output $X^{(n)}$ itself.
When SpecAugment is applied in the latent space, 
augmenting encoder feature~$X^{(n)}$ is a natural approach as prior works~\cite{Wang2021SpecAugmentAH, Devries2017DatasetAI}.
However, augmentation such as masking in the latent feature space tends to cause excessive loss of information in the features, which has a negative impact on training neural networks.
On the other hand, the proposed method can stably obtain the effect of augmentation by applying augmentation only on the conditioning features and remaining the encoder features as the original states. 

Table~\ref{tab:pos-aug2} shows the analysis for the augmentation position to figure out which feature in the latent space are suitable for augmentation.
It can be seen that the direct augmentation of (1) encoder feature~$X^{(n)}$ resulted in performance degradation.
The proposed augmentation for conditioning feature~$C^{(n)}$ was successful in improving performance.

\section{Conclusions}
\label{sec:conclusions}
This paper presented an augmentation method in the intermediate feature space and intermediate token space for Self-conditioned CTC.
Since the intermediate predictions are in the same domain as the output,
it is possible to design augmentations simulating errors that occur upon prediction,
such as insertions, deletions, and substitution errors.
By simulating and correcting these errors through the multiple intermediate augmentations and subsequent intermediate losses, the robustness of encoder model is enhanced.
The experimental comparisons demonstrated that the proposed augmentation methods boost the speech recognition performance of Self-conditioned CTC while keeping the model size lightweight.


\bibliographystyle{IEEEtran}

\bibliography{mybib}

\end{document}